\documentclass[10pt,twocolumn,letterpaper]{article}

\usepackage{cvpr}              

\def\eg{\emph{e.g}\onedot} 
\def\ie{\emph{i.e}\onedot}

\definecolor{cvprblue}{rgb}{0.21,0.49,0.74}
\usepackage[pagebackref,breaklinks,colorlinks,allcolors=cvprblue]{hyperref}
\usepackage{pifont}
\usepackage{multirow}
\usepackage{amsmath}
\usepackage{algorithm}
\usepackage{algpseudocode}
\usepackage{amssymb}    
\usepackage{graphicx}    
\usepackage{booktabs}  
\usepackage{bbding}

\usepackage[capitalize,noabbrev]{cleveref}
\usepackage[table]{xcolor}  
\def\paperID{19275} 
\def\confName{CVPR}
\def\confYear{2026}

\title{EventGait: Towards Robust Gait Recognition with Event Streams}

\author{Senyan Xu$^{1,*}$, Shuai Chen$^{1,}$\footnotemark[1], Chuanfu Shen$^{2,*}$, Kean Liu$^{1}$, Zhijing Sun$^{1}$, Chengzhi Cao$^{1}$, Xueyang Fu$^{1,}$\footnotemark[2]\\
 $^{1}$University of Science and Technology of China\\
 $^{2}$University of Electronic Science and Technology of China\\
{\tt\small \{syxu,shuaic\}@mail.ustc.edu.cn, chuanfu.shen@uestc.edu.cn, \{xyfu\}@ustc.edu.cn}
}

\begin{document}
\maketitle
\footnotetext[1]{Equal contribution.}
\footnotetext[2]{Corresponding author.}
\begin{abstract}
Gait recognition enables non-intrusive, privacy-preserving identification but suffers in uncontrolled environments due to illumination and motion sensitivity of conventional cameras. In this work, we explore gait recognition using event cameras, which offer microsecond temporal resolution and high dynamic range, naturally capturing robust dynamic cues and suppressing static noise. Existing event-based approaches typically aggregate event streams into event images over long time windows, thereby discarding fine-grained motion dynamics critical for gait recognition.
Therefore, we propose \textbf{EventGait}, an end-to-end dual-stream framework that separately models motion and shape while preserving the advantages of events. Our dynamic stream leverages a Mixture of Spiking Experts (MoSE) with diverse neuron constants for robust dynamic perception across complex motion and illumination scenes, while the static stream learns dense shape representations via Cross-modal Structure Alignment (CroSA) with large vision foundation models. 
To address the absence of large-scale event-based gait datasets, we introduce a synthesis pipeline and release two new benchmarks: SUSTech1K-E and CCGR-Mini-E. Extensive experiments have shown that event-based gait recognition not only achieves results comparable to camera-based gait recognition under normal conditions but also significantly outperforms it in low-light scenarios. Our approach sets a new state of the art on both synthesized and real-world event-based gait benchmarks, highlighting the robustness and potential of event-driven gait analysis. The code and datasets are released at https://github.com/QUEAHREN/EventGait.
\end{abstract}

\section{Introduction}
Gait has emerged as a promising biometric for non-intrusive, privacy-preserving human identification. By utilizing their walking pattern, gait recognition offers the advantage of operating from a distance, making it valuable for surveillance~\cite{shen2024comprehensive}, healthcare~\cite{zafra2025health}, and forensics~\cite{nixon2006automatic}. 
However, conventional RGB-based gait recognition~\cite{li2025rethinking,xionghj2024causality} suffers from unconstrained environments. The illumination changes, occlusions, and motion blur often corrupt visual cues, leading to unreliable performance. Recent LiDAR-based gait recognition~\cite{shen2023lidargait,shen2025lidargait++} has achieved greater robustness, but its high cost and energy consumption hinder deployment at scale. 

\begin{figure}[t]
\centering
\includegraphics[width=0.95\columnwidth]{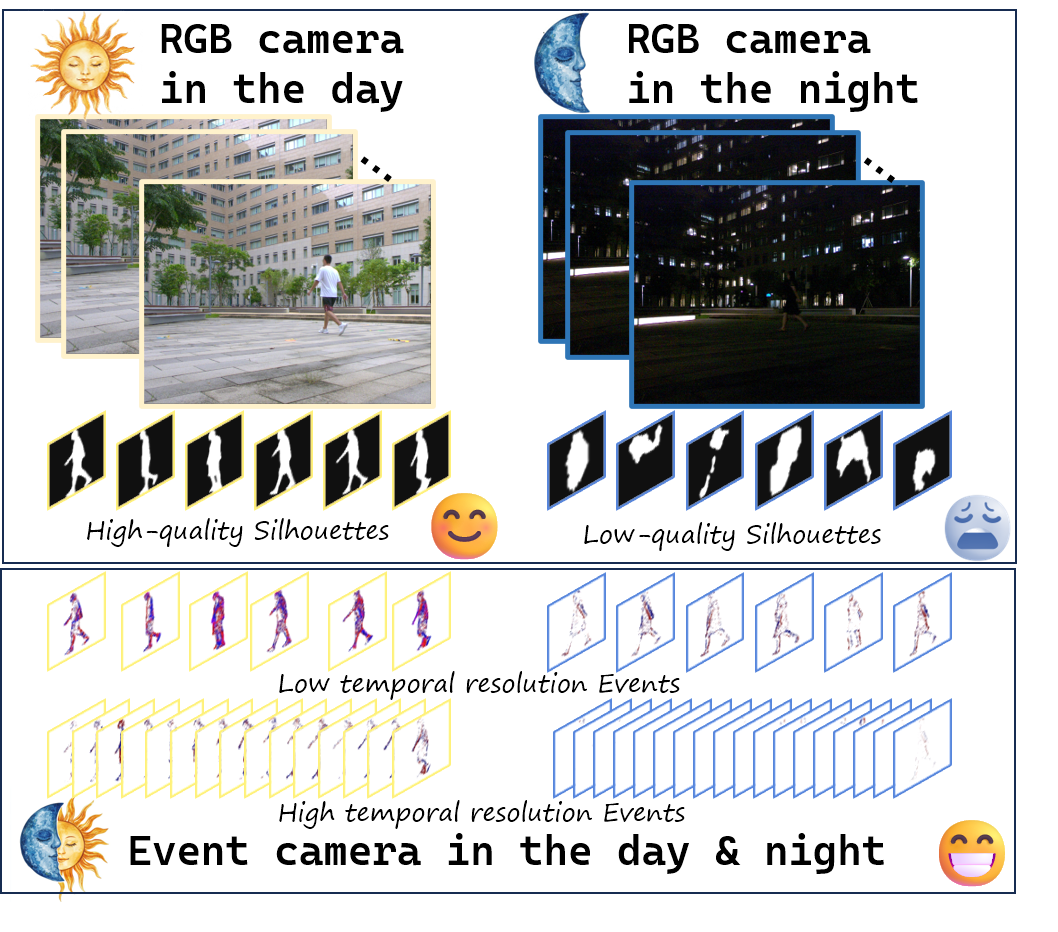}
\vspace{-1em}
\caption{Comparison of RGB and event cameras under day and night conditions. RGB cameras fail to capture usable gait representations in low-light conditions. While event cameras are robust to illumination changes and capture extremely high-temporal-resolution data, preserving spatiotemporal cues in all lighting.}
\label{fig:overview}
\vspace{-1.5em}
\end{figure}

Event cameras, or Dynamic Vision Sensors (DVS)~\cite{Davis346,prophessgen4,samsungdvs}, are bio-inspired sensors that asynchronously record per-pixel brightness changes, providing higher temporal resolution ($<3{\mu}s$) and an extraordinary dynamic range ($>120$\,dB). While conventional cameras have limited temporal resolution ($\approx 30$\, ms) and dynamic range ($<80$\,dB). Event cameras naturally capture rich dynamic motion~\cite{gallego2017event6dof,xu2025motion,xu2024demosaicformer,cao2026learning,fu2024event,sun2025evdm,zhong2025compevent,ge2025eventmamba}, while discarding irrelevant cues such as clothing textures and color. These strengths enable robust perception across challenging lighting conditions, making them suitable for gait recognition. 

With these advantages, the first event-based gait recognition method~\cite{wang2019evgait} has shown the feasibility of using event camera for robust gait recognition. However, prior work~\cite{wang2019evgait,wang2021evgaitpami} has not fully exploited the potential of event data, integrating events over long time windows into grid-based event images. Although these event images can be encoded using standard CNNs~\cite{wang2019evgait} or GNNs~\cite{wang2021evgaitpami}, this temporal voxelization produces sparse representations with low temporal precision. This paradigm is suboptimal for two key reasons: i) it collapses fine temporal resolution, destroying the high-frequency dynamic cues crucial for gait recognition; and ii) it produces overly sparse spatial representations, which conventional deep networks struggle to interpret into dense and discriminative appearance features~\cite{cao2023eventreid}, as shown in \cref{fig:overview}.

In this paper, we revisit event-based gait recognition from a new perspective: robust event-based gait recognition should not just encode spatially static body shape. More importantly, detailed dynamics should be retained from high-temporal-resolution events as well.
Building upon this insight, we propose EventGait, an event-based dual-stream gait recognition framework that separately models motion and shape while preserving the advantages of events.  
The Dynamic Motion stream employs spiking dynamics as basic neurons to model high-frequency motion features directly. 
We further implement a Mixture of Spiking Experts (MoSE), which combines these spiking experts with diverse membrane time constants, enabling robust dynamic perception across complex motion and illumination scenes.
The Static Shape stream enhances the spatial density of structure features through Cross-modal Structure Alignment (CroSA) from a pre-trained vision foundation model (VFM). The VFM, as a teacher, forces the static encoder to learn to obtain dense structural priors from event modality.  

Due to a lack of large-scale datasets for evaluating event-based gait recognition, we establish a synthetic dataset pipeline to simulate event streams from RGB gait videos under different light conditions. 
With the proposed event synthetic pipeline, we construct two large-scale event-based gait datasets, SUSTech1K-E and CCGR-Mini-E, which enrich the diversity of training samples and also serve as strong new benchmarks for evaluation. 
Our main contributions are summarized as follows:
\begin{itemize}
    \item We present EventGait, a dual-stream framework that separately models static shape and dynamic motion for robust event-based gait recognition.
    \item We propose a Mixture of Spiking Experts (MoSE), where experts with distinct spiking neuron time constants enable adaptive motion perception under varying illumination.
    \item We design a Cross-modal Structure Alignment (CroSA) that transfers structural priors from a VFM teacher to empower event representations.
    \item We establish an event synthesis pipeline and construct two large-scale synthetic event gait datasets, SUSTech1K-E and CCGR-Mini-E, to facilitate large-scale event-based gait analysis.
\end{itemize}
Extensive experiments show that EventGait achieves a state-of-the-art performance compared to previous camera-based and event-based methods, across synthesized and real-world benchmarks.

\begin{figure*}[t]
\centering
\includegraphics[width=1.9\columnwidth]{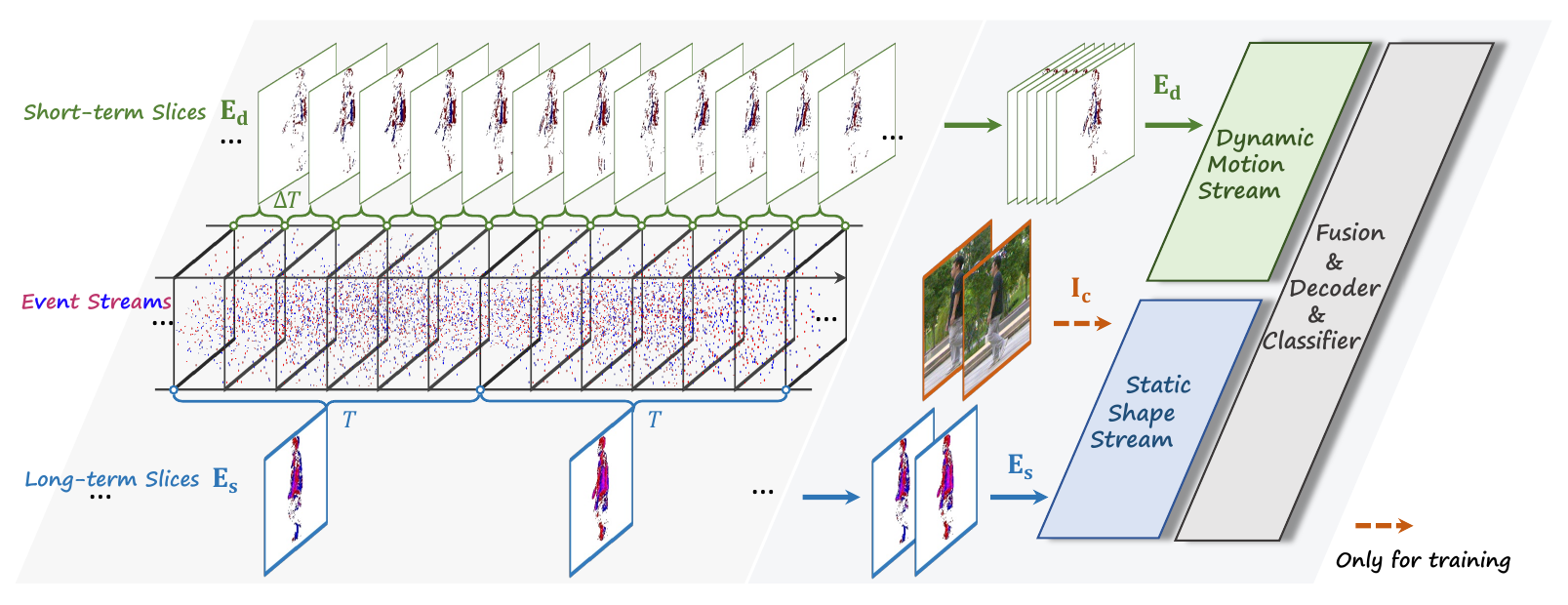}
\vspace{-1em}
\caption{
The workflow of our EventGait.
The detailed architecture of the \textbf{Dynamic Motion Stream} is shown in~\Cref{fig:mose}, while the \textbf{Static Shape Stream} is illustrated in~\Cref{fig:crosa}.
}
\vspace{-1.5em}
\label{fig:eventgait}
\end{figure*}

\section{Related Work}
\subsection{Gait Recognition}
Gait recognition has been extensively researched for decades~\cite{gaitgl,guo2025gait,lin2020gait,xu2023occlusion,li2022multi,shehata2025oumvlpflow}, aiming to identify individuals by leveraging both static shape cues and dynamic motion patterns across diverse modalities, such as silhouettes~\cite{wang2023causal,huangxh2021contextCSTL,xionghj2025gaitdiffusion,wang2024free,huang2024occluded,wang2024qagait,huangxh2023condition,zhou2025exploring,huang2025learning,wang2025gaitc3i}, skeletons~\cite{liao2017pose,fan2023skeletongait,fu2023gpgait,meng2025fastposegait,meng2025seeing,lang2025beyond}, parsings~\cite{wang2023gaitparsing,zheng2023parsing,zou2023cross,zheng2024takes}, RGB imagery~\cite{zhang2019gait,liang2022gaitedge,ye2024biggait,jin2025denoising,wang2025gait,ye2025biggergait}, point clouds~\cite{shen2023gait,wang2023pointgait,guo2024camera,han2024gait,shen2025lidargait++}, and event streams~\cite{wang2019evgait,wang2021evgaitpami}. Traditionally, binary silhouettes have been the dominant modality, effectively retaining body shape while filtering out irrelevant background texture. Similarly, skeleton-based methods have been explored for their explicit structural information, which offers inherent robustness to viewpoint variations. However, both representations are intermediate and heavily dependent on the estimation quality of upstream tasks (\ie, human segmentation and pose estimation). To address this, recent research~\cite{ye2024biggait,fu2024cut,li2021end,li2020end} has shifted toward end-to-end RGB-based methods, achieving state-of-the-art results in both controlled and in-the-wild settings. Nevertheless, RGB imagery remains susceptible to visual ambiguity arising from poor illumination and complex outdoor backgrounds. Consequently, recent studies have adopted 3D sensors to represent gait via point clouds, ensuring robust performance immune to illumination changes. Image-based methods are illumination-sensitive~\cite{dou2022gaitmpl,dou2022metagait,dou2023gaitgci}, and 3D sensors are cost-prohibitive~\cite{shen2023lidargait}. 
This trade-off motivates our exploration of event-based representations. 
Event streams are illumination-robust, offer exceptional dynamic range and high temporal resolution, and are increasingly affordable, making them promising for capturing gait features.

\subsection{Event-based Vision}
Event cameras, or Dynamic Vision Sensors (DVS)~\cite{Davis346,prophessgen4,samsungdvs}, are bio-inspired sensors that operate fundamentally differently from traditional cameras. Instead of capturing frames at a fixed rate, event cameras capture independent pixels that asynchronously trigger \textit{events} in response to a change in logarithmic brightness. 
Event streams offer significant advantages, including high dynamic range, low latency, and a strong suppression of motion blur, making such sensors ideal for several computer vision tasks, from low-level vision~\cite{peng2023learning,peng2024efficient,peng2024towards,peng2024learning,peng2024lightweight,peng2024unveiling,peng2025boosting,peng2025directing,di2025qmambabsr} (\eg, deblurring~\cite{kim2024fevd,kim2025event,li2024fouriermamba,liu2025event,jiang2024rbsformer} and enhancement~\cite{Liang_2024_EvLight,liang2023coherent,liu2025dreamuhd,lu2025evenformer}) to high-level vision (\eg, tracking~\cite{li20243devtrack,messikommer2023data_ev_track}, segmentation~\cite{chen2024segment}, and recognition~\cite{zubic2024state,cao2023eventreid}). Event data is typically encoded either by grouping events into grid-like tensors for CNNs or by processing them asynchronously with point-based networks~\cite{chen2022efficient} or Spiking Neural Networks (SNNs)~\cite{ghosh2009spiking,ge2026fractionalorder}. While event-based gait recognition~\cite{wang2019evgait,wang2021evgaitpami} has proven effective in poor illumination, existing research~\cite {wang2019evgait,wang2021evgaitpami} has two key limitations: (1) it neglects fine-grained temporal dynamics, relying on aggregated event images across long time windows; (2) it was only validated on small-scale datasets ($\sim$100 identities). We address these gaps by proposing a dual-stream network to encode spatiotemporal information and validating our approach on larger scale datasets ($\sim$2000 identities).

\section{EventGait}

In this section, we revisit the event representation and introduce a \textbf{two-scale temporal design} that decomposes the event stream into short- and long-term voxels. 
Then, we detail the \emph{Dynamic Motion Stream} based on the Mixture of Spiking Experts (MoSE) for motion modeling, 
and the \emph{Static Shape Stream} trained with Cross-modal Structure Alignment (CroSA) for dense shape learning. 
Finally, we describe the overall architectural and training objectives.

\subsection{Event Representation}
\label{sec:event_representation}
An event camera asynchronously records per-pixel brightness changes as tuples
$e_i=(x_i, y_i, t_i, p_i)$, 
where $(x_i, y_i)$ are spatial coordinates, $t_i$ is the timestamp, and 
$p_i \!\in\!\{+1,-1\}$ denotes the polarity of the logarithmic intensity change.
An event is triggered when the log-intensity variation at a pixel exceeds a threshold $c$:
\begin{align}
p_i = \mathrm{sg}\!\left(\log \frac{L(x_i,y_i,t_i)}{L(x_i,y_i,t_i')}\right), 
 |\log \frac{L(x_i,y_i,t_i)}{L(x_i,y_i,t_i')}| > c,  
\end{align}

where $\mathrm{sg}(\cdot)$ denotes the sign function.To interface this asynchronous stream with deep networks, we aggregate events within an exposure window $T$ into a temporal event slice.
The interval $T$ is uniformly divided into $K$ bins:
\begin{align} 
\mathbf{E}_p(x,y,k)=
\sum_{e_i\in\mathcal{E}_p}
\max\!\left(0,1-\frac{|t_i-t_k|}{\Delta T}\right),
\end{align}
where $\Delta T=T/K$ denotes the temporal interval of each bin, 
$\mathbf{E}\!\in\!\mathbb{R}^{2\times K\times H\times W}$ represents the voxelized event representation slices, which encodes positive and negative polarities in separate channels, retaining sub-bin temporal precision and motion continuity.

\noindent \textbf{Two-scale temporal design for dual-stream framework.}
Within each long exposure window $T$, our event representation provides $K$ short temporal event slices of width $\Delta T = T / K$.
We exploit this hierarchy through a dual-stream design. As shown in~\cref{fig:eventgait},
the \emph{Dynamic Stream} processes short-term event slices $\mathbf{E}_d$ with small $\Delta T$ to preserve fine-grained temporal fidelity, while the \emph{Static Stream} aggregates events over the entire window $T$ into a long-term event slice $\mathbf{E}_s$, capturing stable spatial shape.

\begin{figure*}[t]
\centering
\includegraphics[width=1.9\columnwidth]{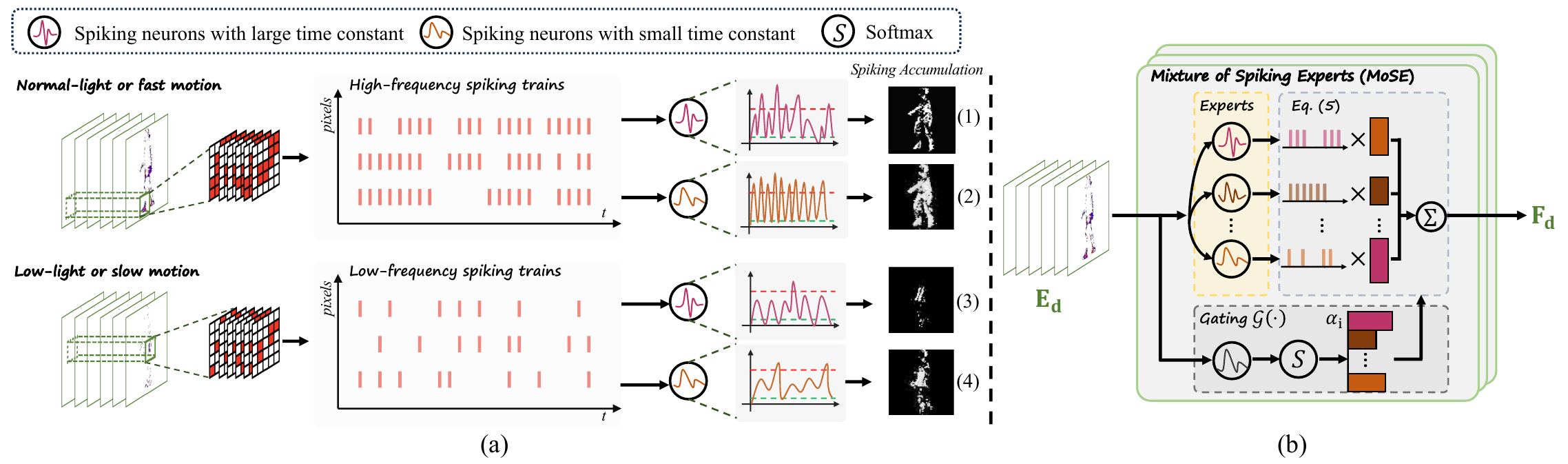}
\vspace{-1em}
\caption{
(a) The simplified schematic of the different spiking neurons' dynamics across complex conditions (\eg, illumination and motion) for intuitive understanding, (b) The details of the Dynamic Motion Stream, which consists of our Mixture of Spiking Experts (MoSE).
}
\vspace{-1.5em}
\label{fig:mose}
\end{figure*}

\begin{figure}[t]
\centering
\includegraphics[width=0.9\columnwidth]{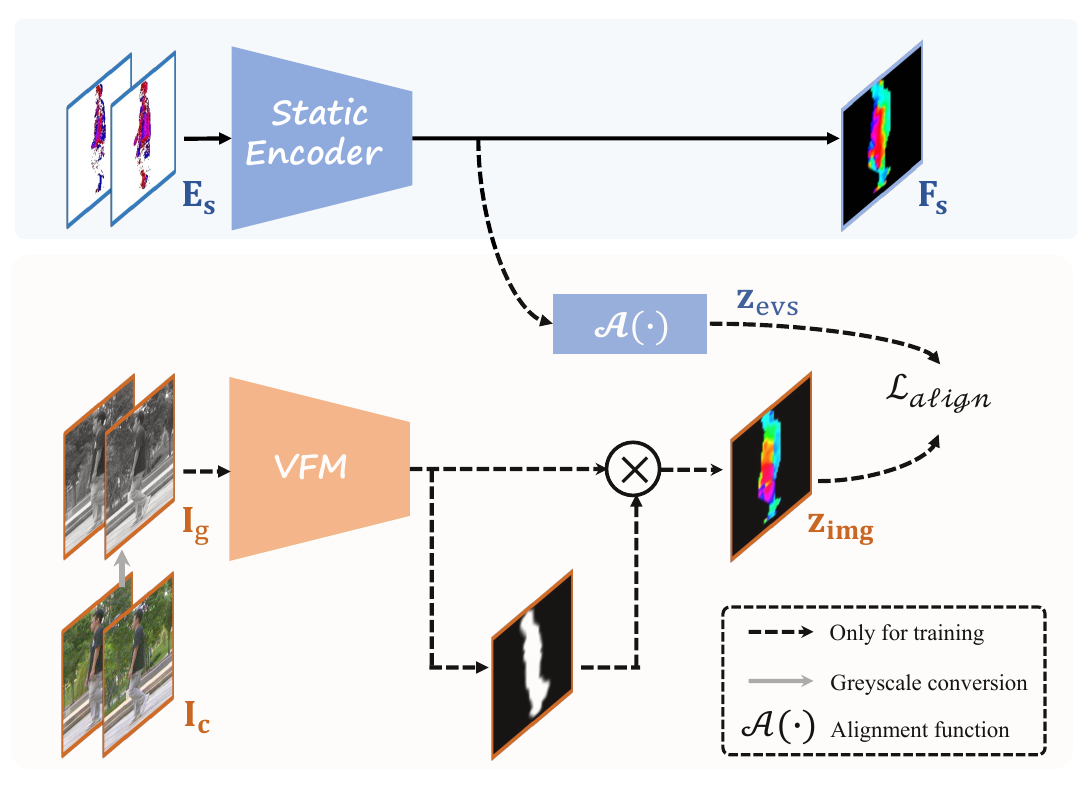}
\vspace{-0.5em}
\caption{
The details of the Static Shape Stream, which is trained with our Cross-modal Structure Alignment (CroSA).
}
\label{fig:crosa}
\end{figure}

\subsection{Mixture of Spiking Experts}
The dynamic motion stream operates on high temporal frequency, and spatially sparse event slices $\mathbf{E}_d$ with high temporal resolution.
Such signals differ from dense frames. Traditional CNNs/RNNs-based methods struggle to capture precise motion features and are incompatible with the inherent sparsity of events.

\noindent \textbf{Modeling as spiking dynamics.} The Spiking Neural Networks (SNNs)~\cite{ghosh2009spiking} integrate sparse spikes over time to update an internal state, providing a naturally suitable mechanism for accumulating dynamic cues from our event representation.
Specifically, the membrane potential $U(t)$ of a Leaky Integrate-and-Fire (LIF) neuron~\cite{tavanaei2019lif} evolves according to:
\begin{equation}
\tau \frac{dU(t)}{dt} = -U(t) + R \cdot I(t),
\quad
S(t) = \Theta(U(t) - U_{th}),
\label{eq:LIF}
\end{equation}
where $I(t)$ is the total synaptic input current, $\tau$ is the membrane time constant controlling the rate of potential decay, $R$ is the membrane resistance, $U_{th}$ is the firing threshold, and $\Theta(\cdot)$ is the Heaviside step function. When $U(t)$ exceeds $U_{th}$, the neuron fires a spike ($S(t)=1$) and resets to a resting potential $U_{reset}$. The synaptic current integrates pre-synaptic spikes from connected neurons:
\begin{equation}
I(t) = \sum_i w_{i} \sum_k \psi(t - t_i^{(k)}),
\label{eq:synaptic}
\end{equation}
where $w_{i}$ is the synaptic weight, $t_i^{(k)}$ is the $k$-th spike time from neuron $i$, and $\psi(\cdot)$ is a temporal kernel that models post-synaptic current decay.

\noindent \textbf{From single unit to mixture of experts.}
The dynamics of the LIF neuron (\cref{eq:LIF}) are critically governed by parameters like the membrane time constant $\tau$, which dictates the neuron's spatial-temporal perception.
As shown in~\cref{fig:mose}(a), single neurons with fixed configurations make them highly specialized, making it challenging to model the diverse signals~\cite{yang2024latency,liu2024seeing} from an event camera across complex motion and illumination.
For instance, a neuron with a short $\tau$ (a "fast" neuron) excels at capturing high-frequency bursts (common in bright scenes or fast motion, see~\cref{fig:mose}(a)(1)), but it fails to integrate sparser signals (common in low-light scenes or slow motion, see~\cref{fig:mose}(a)(3)). 
Besides, a neuron with a large $\tau$ (a ``slow'' neuron) demonstrates robustness in low-light or slow motion conditions by accumulating sparse spikes over a long time (\cref{fig:mose}(a)(4)), but it introduces temporal noise in bright or fast motion conditions (\cref{fig:mose}(a)(2)). 
In real-world scenarios, the coupling of diverse illumination and motion results in far more dynamic spike patterns. Consequently, a neuron unit with a single configuration is insufficient to adapt to complex environments.

Therefore, we argue that a single spiking unit cannot generalize across these varied signal dynamics. Motivated by Mixture of Experts~\cite{masoudnia2014mixture}, we introduce the \textbf{Mixture of Spiking Experts (MoSE)}—an ensemble of spiking sub-neurons specialized across temporal regimes, to improve the robustness of the dynamic streams across complex motion or illumination conditions, as shown in \cref{fig:mose}(b). MoSE consists of $N$ parallel spiking experts $\{\mathcal{E}_1,\dots,\mathcal{E}_N\}$, each initialized with a distinct membrane constant $\tau_i$.
A lightweight spiking gating network $\mathcal{G}(\cdot)$ analyzes the event's dynamic pattern and computes adaptive mixture coefficients $\{\alpha_i\}$ for each expert, producing the final dynamic motion feature:
\begin{align}
   \hat{\mathbf{E}}_t = \sum_{i=1}^{N} \alpha_i\, \mathcal{E}_i(\mathbf{E}_t), 
\end{align}
where $\mathbf{E}_t$ denotes the input event tensor. Experts with small $\tau_i$ specialize in high-frequency, bright-scene motion, while those with larger $\tau_i$ integrate sparse events from darker conditions. We employ MoSE as the basic unit of the dynamic encoder, yielding robust motion representations.

\subsection{Cross-modal Structure Alignment}
While the dynamic stream captures robust short-term motion patterns, gait recognition also requires stable static structural cues that characterize the canonical human shape.
As discussed in~\cref{sec:event_representation}, we employ long-term integration $\mathbf{E}_s$ to obtain full spatial statistics, but it also remains a sparse structure. 
It is difficult for an encoder to learn complex, dense human shape priors from this sparse input without additional guidance.

To overcome this, we propose \textbf{Cross-modal Structure Alignment (CroSA)}, a strategy to distill rich structural priors from a pre-trained Vision Foundation Model (VFM)~\cite{oquab2023dinov2} into our static stream, as presented in~\cref{fig:crosa}.
Our goal is to teach our event-based static encoder ($\mathcal{F}_{\text{student}}$), implemented as CNNs, to understand human structure and shape.
We apply a frozen DINOv2~\cite{oquab2023dinov2} as the teacher $\mathcal{F}_{\text{teacher}}$, as it is renowned for its strong grasp of fine-grained structural information compared to others~\cite{clip}.

Given a synchronized RGB frame $\mathbf{I}_{c}$, we convert it to an intensity image $\mathbf{I}_{g} = \mathcal{I}(\mathbf{I}_{c})$  via a grayscale operation $\mathcal{I}(\cdot)$ to avoid irrelevant color disturbance from RGB~\cite{ye2024biggait}, and then feed it to the teacher encoder:
\begin{align}
\mathbf{z}_{\text{img}} = 
\mathcal{F}_{\text{teacher}}(\mathbf{I}_{\text{g}}), \quad\mathbf{z}_{\text{evs}} = \mathcal{A}(\mathcal{F}_{\text{student}}(\mathbf{E}_s)),
\end{align}
where $\mathcal{A}(\cdot)$ is an alignment convolution layer to project the event feature. The alignment loss is to minimize the distance between $\mathbf{z}_{\text{img}}$ and $\mathbf{z}_{\text{evs}}$:
\begin{align}
\mathcal{L}_{\text{align}} = 
\|\mathbf{z}_{\text{evs}} - \mathbf{z}_{\text{img}}\|_2^2,
\end{align} 
where $\ell_2$ denotes the Euclidean distance.

\subsection{Overall Architecture}
Our proposed \textbf{EventGait} follows a dual-stream design, as shown in~\cref{fig:eventgait}. 
The \emph{Dynamic Stream} is implemented using the MoSE as basic learning units, producing a dynamic representation $\mathbf{F}_{\text{d}}$ from the short-term slices $\mathbf{E}_d$. 
In parallel, the \emph{Static Stream} employs a CNN-based encoder trained with CroSA, producing a dense shape feature $\mathbf{F}_{\text{s}}$ from long-term events $\mathbf{E}_s$. The two representations are fused through a conventional fusion module $\Phi(\cdot)$:
\begin{align}
\mathbf{F}_{\text{gait}} = \Phi([\mathbf{F}_{\text{s}}; \mathbf{F}_{\text{d}}]),
\end{align}
forming a unified gait descriptor integrating event-native sensing and dense-vision priors.
The fused embedding is then fed into the downstream recognition module. We also utilize cross-entropy loss $\mathcal{L}_{\text{ce}}$ and the triplet loss $\mathcal{L}_{\text{tri}}$, and the overall training objective is formulated as:
\begin{align}
\mathcal{L}_{\text{total}} 
= \mathcal{L}_{\text{ce}} 
+ \mathcal{L}_{\text{tri}} 
+ \lambda_{\text{d}} \mathcal{L}_{\text{align}},
\end{align}
where $\lambda_{\text{d}}$ balances the contribution of the cross-modal alignment term.

\section{Construction of Event-based Datasets}
Existing gait recognition datasets are predominantly captured by conventional cameras, leaving a notable void in large-scale, event-based benchmarks. To facilitate research in event-based gait recognition, we construct two new datasets with realistic synthetic events from intensity frames. In this section, we detail our event-based benchmark synthesis pipeline. We converted two popular existing datasets into their event-based counterparts (\ie CCGR-Mini-E, and SUSTech1K-E).

\subsection{Event Synthetic Pipeline}
Prior research~\cite{cao2023eventreid,kim2024fevd,kim2025event,Liang_2024_EvLight} has demonstrated that synthetic DVS events from standard video frames can enhance model generalization under uncontrolled lighting conditions. Building on this, we adopt a similar paradigm to facilitate event-based gait recognition. We construct a synthesis pipeline to convert RGB videos from existing datasets into event gait streams. This pipeline first performs a frame interpolation network to generate an intermediate sequence. Then, following the v2e toolbox~\cite{hu2021v2e}, by adjusting the setting (\eg, cut-off frequency, temporal noise level), we simulate event streams across a wide range of illumination conditions, following the logarithmic temporal evolution of pixel intensities. Next, we accumulate raw events within the temporal windows to obtain synchronous event slices. Finally, a human detector is applied to the RGB sequences to extract bounding boxes, which are temporally interpolated and then used to crop these pre-accumulated event slices, forming our final discrete event representation.

\subsection{Synthetic Event-based Datasets}
With our proposed event synthetic pipeline, we construct two large-scale event-based gait datasets from two widely-used RGB-based datasets. 

\noindent \textbf{CCGR-Mini-E} is synthesized from the CCGR-Mini dataset~\cite{zou2023cross} using our synthetic pipeline. CCGR-Mini-E is distinguished by its significant scale and diversity. It contains 47,884 sequences from 970 subjects, and each subject has 53 distinct covariates and 33 different views. To the best of our knowledge, CCGR-Mini-E is the largest and most covariate-rich event-based gait dataset available.

\begin{table*}[!t]
\centering
\caption{Within-domain Evaluation on SUSTech1K~\cite{shen2023lidargait} and SUSTech1K-E (ours).}
\vspace{-0.6em}
\resizebox{1.95\columnwidth}{!}{ 
\begin{tabular}{c|c|c|c|ccccccccc} 
\toprule
\multicolumn{1}{c|}{\multirow{2}{*}{Input}} & \multirow{2}{*}{Model} & \multirow{2}{*}{Venue} & \multirow{2}{*}{Params} & \multicolumn{9}{c}{Rank-1} \\ 
\cmidrule{5-13} 
\multicolumn{1}{c|}{}& & & & NM & BG & CL & CR & UB & UN & OC & NT & Overall \\ 
\midrule

\multicolumn{13}{c}{\cellcolor{gray!8}\textbf{SUSTech1K}} \\ 
\midrule
 \multirow{2}{*}{PCs}
 & HMRNet~\cite{han2024gait} & MM2024 & - & 92.7 & 92.3 & \underline{79.6} & 90.3 & 83.1 & \underline{95.2} & 86.2 & \underline{90.4} & 90.2 \\ 
 & LidarGait++~\cite{shen2025lidargait++} & CVPR2025 & 4.4M & \textbf{94.2} & \textbf{93.9} & \textbf{79.7} & \underline{92.4} & \underline{91.5} & \textbf{96.6} & 91.9 & \textbf{92.2} & \underline{92.7} \\ 

\midrule
\multirow{5}{*}{Sils} 
 &GaitSet~\cite{Chao2019} & AAAI2019 & 2.6M & 69.1 & 68.3 & 37.4 & 65.0 & 63.1 & 61.0 & 67.2 & 23.0 & 65.0 \\
 & GaitPart~\cite{fan2020gaitpart} & CVPR2019 & 1.2M & 62.2 & 62.8 & 33.1 & 59.5 & 57.3 & 54.9 & 57.2 & 21.8 & 59.2 \\ 
 & GaitBase~\cite{fan2022opengait} & CVPR2023 & 8.0M & 81.5 & 77.5 & 49.6 & 75.8 & 75.6 & 76.7 & 81.4 & 25.9 & 76.1 \\ 
 & DeepGaitV2~\cite{fan2025opengait} & TPAMI2025 & 9.1M & 87.4 & 84.1 & 53.4 & 81.3 & 86.1 & 84.8 & 88.5 & 28.8 & 82.3 \\ 
 & GaitLLM-10~\cite{yang2025bridging} & CVPR2025 & 23.4M & 88.2 & 86.3 & 59.7 & 83.5 & 88.8 & 86.9 & 90.5 & 28.8 & 84.5 \\
 & Origin-S~\cite{huang2025learning} & ICCV2025 & - & 91.4 & 88.1 & 64.8 & 86.0 & 89.8 & 88.9 & 92.8 & 29.6 & 86.9 \\
& $\alpha$-Gait-S~\cite{huangvocabulary} & NIPS2025 & - & 91.1 & 87.2 & 64.0 & 85.3 & 89.5 & 88.8 & 92.7 & 28.2 & 86.3 \\

\midrule

  Sils+Skeleton
 & SkeletonGait++~\cite{fan2023skeletongait} & AAAI2024 & 9.1M & 85.1 & 82.9 & 46.6 & 81.9 & 80.8 & 82.5 & 86.2 & 47.5 & 81.3  \\ 

\midrule

 Sils+Parsing+Flow
 & MultiGait++~\cite{jin2024exploring} & AAAI2025 & 12.1M & 92.0 & 89.4 & 50.4 & 87.6 & 89.7 & 89.1 & 93.4 & 45.1 & 87.4 \\ 

\midrule

 Sils+Depth Map
 & DepthGait~\cite{li2025depthgait} & MM2025 & - & \underline{93.5} & 88.0 & 56.8 & 87.4 & 90.0 & 88.5 & \underline{95.2} & 38.4 & 87.6 \\ 

\midrule
\multicolumn{13}{c}{\cellcolor{gray!20}\textbf{SUSTech1K-E (Ours)}} \\ 
\midrule


 \multirow{3}{*}{Event}
 & EVGait~\cite{wang2019evgait} & CVPR2019 & 45.2M & 55.0 & 70.8 & 76.7 & 67.8 & 58.8 & 44.8 & 58.5 & 78.7 & 65.4 \\ 
 & GaitBase$^e$~\cite{fan2022opengait} & CVPR2023 & 8.0M & 66.7 & 62.0 & 40.6 & 63.8 & 65.2 & 53.0 & 61.4 & 59.2 & 63.1 \\ 
 & EventGait & \textbf{Ours} & \textbf{4.6M} & 92.5 & \underline{93.3} & 78.1 & \textbf{93.3} & \textbf{92.8} & 89.7 & \textbf{96.9} & 84.8 & \textbf{92.8} \\

\bottomrule
\end{tabular}
}
\vspace{-0.6em}
\label{tab:sustech1k_withindomain}
\end{table*}

\noindent \textbf{SUSTech1K-E} is synthesized from the SUSTech1K dataset~\cite{shen2023lidargait} using our synthetic pipeline. Its primary distinction is its multi-modal nature, offering paired data across RGB, silhouette, skeleton, point cloud, and our proposed event modalities. The dataset comprises 25,239 sequences from 1,050 identities. SUSTech1K-E retains diverse challenging conditions, including diverse viewpoints, occlusions, lighting, and clothing conditions. It is specifically designed to facilitate analysis and evaluation of modality effectiveness. 

\subsection{Real-world Event-based Datasets}
\noindent \textbf{DVS128-Gait}~\cite{wang2021evgaitpami} is a real-world dataset collected using a DVS128 event camera. It contains 4,000 event streams from 20 subjects, captured during the daytime. The sensor was set at approximately 90 degrees to the walking direction. 

\noindent \textbf{EV-CASIA-B} is a playback-based event dataset derived from CASIA-B dataset~\cite{yu2006framework}. It contains 124 subjects, each providing 66 sequences from 11 distinct view angles. EV-CASIA-B was converted by playing the original RGB videos on a 23-inch Dell monitor ($1920\times1080\,@\,60Hz$) and re-recording them using a DVS128 event camera.

\section{Experiments}
 
\subsection{Implementation Details}
(1) All models are trained on 8 RTX 3090 GPUs.
(2) The SGD optimizer with an initial learning rate of 0.1 and weight decay of 0.0005 is utilized;
(3) All images are processed with Pad-and-Resize~\cite{ye2024biggait} to keep body proportion and resized to 64×64.
(4) More details are provided in the \textbf{Appendix}.

\begin{table}[!t]
\centering
\caption{Within-domain Evaluation on CCGR-Mini~\cite{zou2023cross} / CCGR-Mini-E (Ours) and CASIA-B*~\cite{yu2006framework} / EV-CASIA-B~\cite{wang2019evgait}.}
\vspace{-0.6em}
\resizebox{1\columnwidth}{!}{ 
\begin{tabular}{c|c|c|ccc|c} 
\toprule
\multirow{2}{*}{Input} & \multirow{2}{*}{Model} & \multirow{2}{*}{Venue} &\multicolumn{3}{c|}{\cellcolor{gray!8}\textbf{CCGR-Mini}} & \multicolumn{1}{c}{\cellcolor{gray!10}\textbf{CASIA-B*}} \\ 
\cmidrule{4-7} 
& & & Rank-1 & mAP & mINP & NM \\ 
\midrule
\multirow{4}{*}{\begin{tabular}[c]{@{}c@{}}Sils\end{tabular}} 
& GaitSet~\cite{Chao2019} & AAAI2019 & 13.8 & 15.4 & 5.8 & 92.3 \\
& GaitPart~\cite{fan2020gaitpart} & CVPR2019 & 8.0 & 10.1 & 3.5 & 96.2 \\  
& GaitBase~\cite{fan2022opengait} & CVPR2023 & 27.0 & 24.9 & 9.7 & \underline{96.5} \\ 
& DeepGaitV2~\cite{fan2025opengait} & TPAMI2025 & \underline{39.4} & \underline{36.0} & \underline{16.8} & 94.3 \\  
\midrule
\multirow{3}{*}{\begin{tabular}[c]{@{}c@{}} Skeleton\end{tabular}} 
& GaitGraph2~\cite{teepe2022towards} & CVPRW2022 & 1.2 & 2.4 & 0.6 & 80.3 \\  
& GPGait~\cite{fu2023gpgait} & ICCV2023 & 5.0 & 6.8 & 1.8 & 93.6 \\  
& CAG~\cite{huangxh2023condition} & TIP2023 &  & - &  & 96.4 \\  
\midrule
\multirow{1}{*}{\begin{tabular}[c]{@{}c@{}} Sils+Skeleton\end{tabular}} 
& BiFusion~\cite{peng2024learning} & MTA2023 &  & - &  & 93.0 \\  
\midrule
\multicolumn{3}{c|}{\cellcolor{gray!20}} &\multicolumn{3}{c|}{\cellcolor{gray!20}\textbf{CCGR-Mini-E (Ours)}} & \multicolumn{1}{c}{\cellcolor{gray!20}\textbf{EV-CASIA-B}} \\ 

\midrule
\multirow{3}{*}{\begin{tabular}[c]{@{}c@{}}Event\end{tabular}} 
& EVGait~\cite{wang2019evgait} & CVPR2019  &  & - &  & 89.9 \\ 
& GaitBase$^e$~\cite{fan2022opengait} & CVPR2023 & 9.7 & 10.7 & 4.2 & 94.8 \\ 
& \textbf{EventGait} &  \textbf{Ours} & \textbf{40.3} & \textbf{38.7} & \textbf{25.5} & \textbf{96.7} \\
\bottomrule
\end{tabular}
}
\vspace{-1.6em}
\label{tab:other_indomain}
\end{table}

\begin{table*}[!t]
\centering
\caption{Cross-domain Evaluation.} 
\vspace{-0.3em}
\resizebox{1.8\columnwidth}{!}{
\begin{tabular}{c|c|cccc|cccc|ccc}
\toprule
\multirow{2}{*}{Modality} & \multirow{2}{*}{Method} & \multicolumn{4}{c|}{CCGR-Mini $\rightarrow$ \colorbox{gray!5}{SUSTech1K}} & \multicolumn{4}{c|}{CCGR-Mini $\rightarrow$ \colorbox{gray!30}{Low-light SUSTech1K}} & \multicolumn{3}{l}{SUSTech1K $\rightarrow$ CCGR-Mini} \\ \cmidrule{3-13}
                          &                         & NM         & CL         & NT         & Overall         & NM         & CL         & NT         & Overall         & Rank-1             & mAP            & mINP            \\ \midrule
\multirow{3}{*}{Sils.}    & GaitSet                 & 23.2       & 13.7       & 13.9       & 23.7            & 9.4       & 5.4       & 2.8       & 8.3            &  3.7                  &4.4                &\textbf{1.4}                 \\
                          & GaitBase                & \textbf{49.0}       & 29.0        & 21.8      & 48.1            & 22.2       & 9.7       & 3.0       & 18.5            &   3.2                 &3.8                &1.1                 \\
                          & DeepGaitV2              &  47.8          & \textbf{30.3}           & 21.9           & \textbf{48.2}                &  \textbf{23.3}          &  10.2          & 3.2           & 18.7                & 3.2                   & 3.8               & 1.1                \\ \midrule
\multirow{2}{*}{Event}    
                          & GaitBase$^e$            & 26.6 & 12.8      & 30.2       & 23.0            & 13.9       & 7.4       & 18.8       & 12.4            & 1.3               & 2.1           & 0.7             \\
                          & EventGait               & 45.3       & 28.7       & \textbf{51.4}       & 45.5                & 22.0        & \textbf{14.4}       & \textbf{27.0}       & \textbf{20.7}            & \textbf{3.7}                & \textbf{4.4}            & 1.3            \\ \bottomrule
\end{tabular}
}
\vspace{-1.em}
\label{tab:crossdomain}
\end{table*}

\begin{table}[!t]
\centering
\caption{Cross-Illumination Evaluation.} 
\vspace{-0.3em}
\resizebox{1.0\columnwidth}{!}{
\begin{tabular}{c|c|cccccccc}
\toprule
\multirow{3}{*}{Modality} & \multirow{3}{*}{Methods} & \multicolumn{8}{c}{SUSTech1K-E}                                                                                          \\ \cmidrule{3-10} 
                          &                          & \multicolumn{4}{c|}{\cellcolor{gray!5}\textbf{Normal Light}}  & \multicolumn{4}{c}{\cellcolor{gray!30}\textbf{Low Light}}                                  \\
                          \cmidrule(lr){3-6}  \cmidrule(lr){7-10} 
                          &                          & \multicolumn{1}{c}{NM} & CL & NT & \multicolumn{1}{c|}{Overall} & \multicolumn{1}{c}{NM} & CL& NT & Overall \\ 
                          \midrule
\multirow{2}{*}{Sils}     & GaitSet                  & 69.1                       &37.4 & 23.0   & \multicolumn{1}{c|}{65.0}        &43.7                       &19.7  & 3.0  &31.7                              \\
                          & GaitBase                 &  81.1                      &48.4 & 25.9  & \multicolumn{1}{c|}{76.1}        &54.4                      &26.0  & 3.5  &41.5                              \\ \midrule
\multirow{2}{*}{Event}    & GaitBase$^e$                 & 66.7                  & 40.6 & 59.2 & \multicolumn{1}{c|}{63.1}        & 24.8              & 13.5 & 33.5 & 23.6                         \\
                          & EventGait              &  \textbf{92.5}                  & \textbf{78.1} & \textbf{84.4}& \multicolumn{1}{c|}{\textbf{92.8}}        & \textbf{83.8}                   & \textbf{58.3} &\textbf{70.5} & \textbf{83.2}                         \\
                          \bottomrule
\end{tabular}
}
\label{tab:light}
\vspace{-0.6em}
\end{table}

\subsection{Main Results}
\noindent \textbf{Within-domain Evaluation}.
Our EventGait is mainly compared with camera-based approaches~\cite{Chao2019,fan2020gaitpart,fan2022opengait,fan2025opengait,yang2025bridging,huang2025learning,huangvocabulary,fan2023skeletongait,jin2024exploring,li2025depthgait} and LiDAR-based approaches~\cite{shen2023lidargait}.

\Cref{tab:sustech1k_withindomain} shows that our EventGait consistently outperforms \textbf{\textit{camera-based methods}} on the SUSTech1K-E dataset, achieving a \textbf{+5.2\%} gain under overall conditions. This advantage becomes particularly significant in difficult conditions, since we observe performance boosts of \textbf{+18.4\%} for cloth-changing (CL) and a remarkable \textbf{+37.3\%} at night (NT). These results demonstrate that event streams enable a more robust gait representation, less affected by appearance variations and illumination changes.
Compared with \textbf{\textit{point-based methods}}, EventGait achieves competitive results across all metrics and even surpasses the state-of-the-art LidarGait++~\cite{shen2025lidargait++} in overall accuracy. Notably, EventGait attains this performance using an event camera that is significantly more affordable than a LiDAR sensor (USD~1.5K per event camera vs. USD~75K per LiDAR sensor), 
showing the strong potential of event-based gait recognition for large-scale deployment.
However, existing \textbf{\textit{event-based methods}} have struggled, often underperforming both camera- and LiDAR-based approaches. For example, GaitBase$^e$ with event streams as input suffers a 13.0\% accuracy drop compared to GaitBase using silhouettes. We attribute this performance gap to two main factors: (1) event streams are spatially sparse and lack spatial connectivity, and (2) standard CNNs struggle to effectively capture the temporal dynamics from event data.
To address these limitations, EventGait introduces a \textit{static shape stream} to recover structural cues and a \textit{dynamic motion stream} leveraging spiking neurons to efficiently model high-frequency motion information from events. Experiments demonstrate that EventGait improves performance by \textbf{+16.7\%} over the silhouette-based GaitBase, validating the effectiveness of our approach for event-based gait recognition.

Table~\ref{tab:other_indomain} presents the within-domain evaluation results on CCGR-Mini-E and EV-CASIA-B.
Our proposed EventGait consistently outperforms existing state-of-the-art methods across all modalities, demonstrating the effectiveness of EventGait and rich spatiotemporal information within event streams. Owing to their high temporal resolution and emphasis on motion-relevant pixels, event streams offer more discriminative cues that naturally align with the nature of gait recognition.

\noindent \textbf{Cross-domain Evaluation}. 
As shown in~\Cref{tab:crossdomain}, we evaluate the cross-domain performance of EventGait.
Although EventGait achieves slightly suboptimal results compared to existing state-of-the-art silhouette-based methods when trained on CCGR-Mini and tested on SUSTech1K, it achieves \textbf{51.4\%} accuracy in the night (NT) scenario, outperforming the best silhouette-based method by a large margin of \textbf{+29.5\%}. This demonstrates strong robustness of EventGait under low-light conditions. Moreover, EventGait significantly narrows the gap between camera- and event-based approaches. While previous event-based methods achieved only \textbf{23.0\%} rank-1 accuracy, our EventGait reaches \textbf{45.5\%}, approaching the \textbf{48.2\%} obtained by DeepGaitV2 using silhouettes. When trained on SUSTech1K and tested on CCGR-Mini, EventGait surpasses silhouette-based methods in both rank-1 and mAP, highlighting its superior cross-domain ability and validating the effectiveness of our dual-stream static-dynamic design for event stream. Furthermore, when trained on CCGR-Mini and tested on SUSTech1K under poor illumination, EventGait consistently outperforms all other methods across all metrics, particularly at night, where it surpasses all camera-based approaches by a significant margin of \textbf{+23.8\%}. This confirms that the event modality remains highly robust to complex illumination variations, even in cross-domain scenarios.

\begin{table}[!t]
\centering
\caption{Within-domain Evaluation on realistic DVS128~\cite{wang2021evgaitpami}} 
\vspace{-0.3em}
\resizebox{0.8\columnwidth}{!}{
\begin{tabular}{cccc}
\toprule
DVS128          & EV-Gait & GaitBase$^e$ & EventGait (Ours) \\
\midrule
Rank-1 Accuracy &  81.8       &   74.4           &  \textbf{87.4}         \\
\bottomrule
\end{tabular}
}
\vspace{-0.6em}
\label{tab:indomain_dvs128}
\end{table}

\noindent \textbf{Cross-illumination Evaluation}. 
As shown in \Cref{tab:light}, we evaluate EventGait on the SUSTech1K-E dataset under varying illumination.
EventGait achieves a state-of-the-art \textbf{92.8\%} overall rank-1 accuracy under normal lighting and maintains \textbf{83.2\%} under low lighting, with only a \textbf{9.6\%} drop.
In contrast, silhouette-based methods experience severe degradation due to reduced visual contrast and structural clarity.
These results highlight the inherent advantages of the event modality. The high temporal resolution and lighting robustness enable robust gait recognition under unconstrained environments. While our dual-stream design further compensates for spatial sparsity, ensuring reliable gait recognition even under challenging scenarios.

\begin{figure}[t]
\centering
\includegraphics[width=1.0\columnwidth]{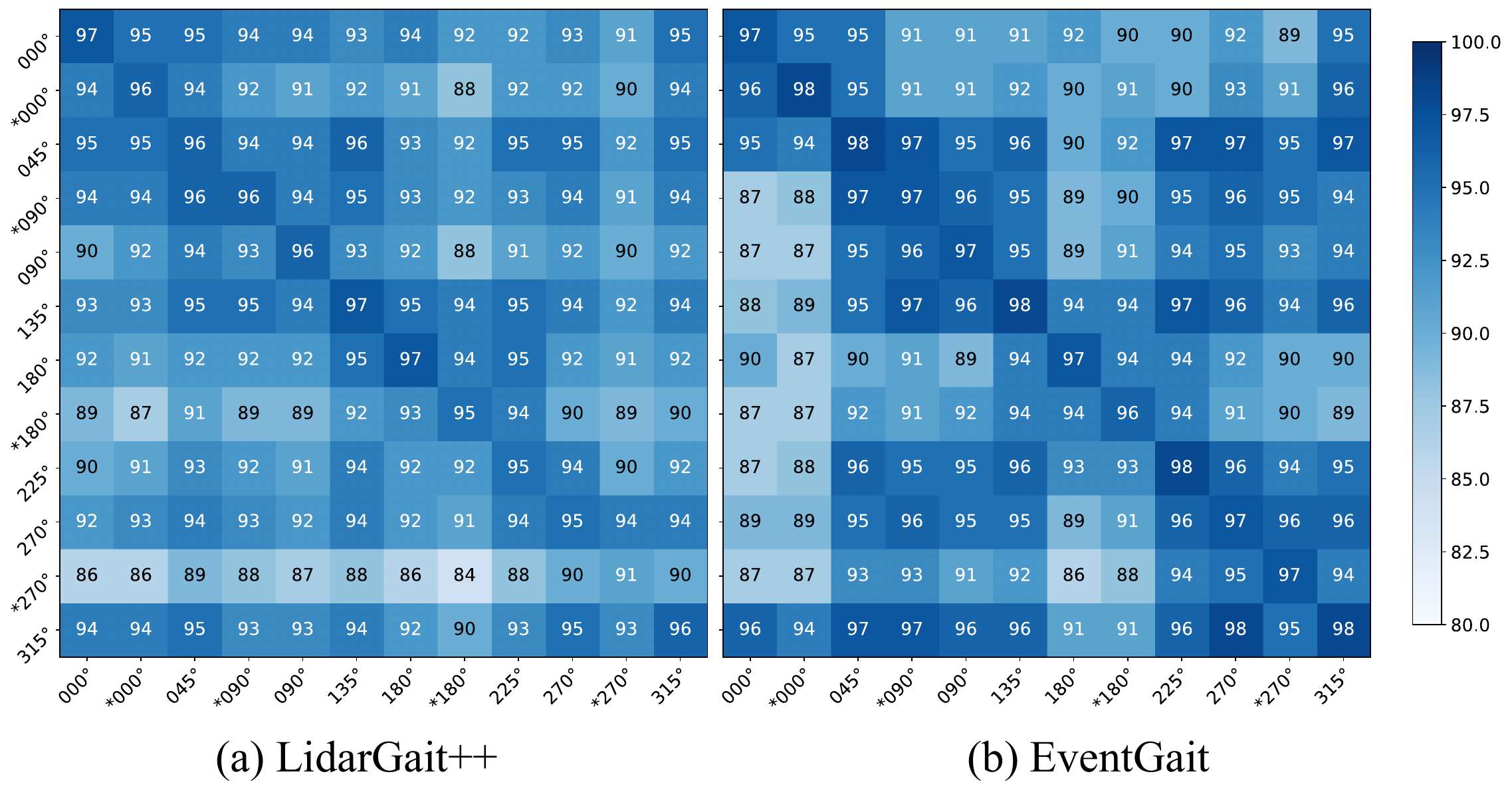}
\caption{
Cross-view performance comparison between LidarGait++~\cite{shen2025lidargait++} and EventGait (ours).
}
\label{fig:comparison_multiview}
\vspace{-0.6em}
\end{figure}

\noindent \textbf{Cross-view Evaluation.} 
As shown in~\Cref{fig:comparison_multiview}, our EventGait achieves performance highly comparable to the SOTA LiDAR-based method~\cite{shen2025lidargait++}, and even outperforms it in several views. This is noteworthy as our 2D event-based method competes directly with 3D LiDAR-based method. We attribute this to the dense spatio-temporal dynamics from events, which offer a richer motion representation than geometrically sparse LiDAR point clouds. Unlike silhouette-based methods that generalize poorly across views~\cite{shen2023lidargait}, EventGait highlights the strong potential of event-based representations for view-invariant recognition.

\noindent \textbf{Real-world Evaluation.} 
We also evaluate our EventGait and compare it with existing state-of-the-art methods on the realistic DVS128-Gait~\cite{wang2021evgaitpami} dataset. As illustrated in~\Cref{tab:indomain_dvs128}, EventGait performs consistently better than prior works, validating the effectiveness of our model on both synthetic and realistic benchmarks.

\begin{table}[!t]
\centering
\caption{Ablation on the static and dynamic streams of EventGait.}
\centering
\resizebox{\columnwidth}{!}{
\begin{tabular}{cc|cccc}
\toprule
\multirow{2}{*}{\begin{tabular}[c]{@{}c@{}}Static Shape \\ Stream \end{tabular}} & \multirow{2}{*}{\begin{tabular}[c]{@{}c@{}}Dynamic Motion\\ Stream\end{tabular}} & \multicolumn{4}{c}{SUSTech1K-E} \\ \cmidrule{3-6} 
                                 &                                      & NM   & CL   & NT   & Overall  \\
                                                                                \midrule
\checkmark                       &   \ding{55}                          & 82.6 & 61.6 & 76.9 &   82.0 \\
\ding{55}                        &    \checkmark                        & 74.5 & 52.0 & 71.7 &   72.4  \\
\checkmark                       &    \checkmark                        & \textbf{92.5} & \textbf{78.1} & \textbf{84.8} &   \textbf{92.8} \\
                                                                                \bottomrule
\end{tabular}
}
\label{tab:branch}
\end{table}

\begin{table}[!t]
\centering
\caption{Ablation of objectives and weights in Cross-modal Structure Alignment.}
\resizebox{\columnwidth}{!}{
\begin{tabular}{c|c|c|cccc}
\toprule
\multirow{2}{*}{Idx} & \multirow{2}{*}{\begin{tabular}[c]{@{}c@{}}Objective \\ in eq. 7\end{tabular}} & \multirow{2}{*}{\begin{tabular}[c]{@{}c@{}}Weight $\lambda_{\text{d}}$\\ in eq. 9 \end{tabular}} & \multicolumn{4}{c}{SUSTech1K-E} \\ \cmidrule{4-7} 
                     &                            &                         & NM   & CL   & NT   & Overall  \\ \midrule
(a)                  & w/o                        & -                       & 87.6 & 71.5 & 78.8 & 87.4     \\
(b)                  & cosine                     & 0.2                     & 88.8 & 70.9 & 81.9 & 89.1     \\
(c)                  & $l_2$                      & 0.05                    & 89.4 & 73.7 & 83.3 & 90.5     \\
(d)                  & $l_2$                      & 0.2                     & \textbf{92.5} & \textbf{78.1} & \textbf{84.8} & \textbf{92.8}     \\
(e)                  & $l_2$                      & 0.5                     & 90.1 & 69.5 & 81.0 & 89.7     \\
\bottomrule
\end{tabular}
}
\label{tab:dinoloss}
\end{table}

\subsection{Ablation Study}
Ablation studies are conducted on SUSTech1K-E~\cite{shen2023lidargait}. 

\noindent \textbf{Static and Dynamic Streams.}
\Cref{tab:branch} highlights the complementary roles of static and dynamic streams in EventGait. 
Removing either stream leads to a noticeable drop in performance, validating the necessity of modeling both within-frame appearance and cross-frame motion. 
Notably, relying solely on the dynamic stream results in suboptimal accuracy, as motion cues alone fail to capture sufficient structural detail, which is consistent with observations in prior work~\cite{jin2025denoising}. 
These results underscore that, even for event data, integrating static and dynamic representations is crucial for capturing complete gait semantics under complex conditions. 
The dual-stream design of EventGait effectively leverages the strengths of event data to extract both dynamic motion patterns and static structural cues, leading to more comprehensive and robust gait representations.

\noindent \textbf{Number of experts in MoSE.}
As shown in~\Cref{tab:experts}, we systematically evaluate the impact of the number of experts in the MoSE module under both normal and low-light conditions. 
Using a single expert (i.e., a standard SNN unit) results in the lowest performance, indicating limited adaptability to diverse motion and illumination scenarios. 
Increasing the number of experts to 3 consistently improves performance across all settings. 
While using 4 experts yields slightly better scores on a few metrics, the overall gains over 3 experts are marginal. Considering the trade-off between accuracy and efficiency, we adopt 3 experts as the default configuration. 
This validates the effectiveness of MoSE in dynamic motion modeling.

\noindent \textbf{About Cross-modal Structure Alignment.}
\Cref{tab:dinoloss} shows the impact of different alignment objectives and weights on performance.
Comparing (a) and (d), introducing our Cross-modal Structure Alignment consistently improves all metrics, indicating its effectiveness in enhancing the structural representation of event data.
From (b) and (c), although cosine provides directional constraints, it lacks fine-grained guidance, leading to performance degradation.
As seen in (c)–(e), $l_2$ achieves the best result when the weight is moderate ($\lambda_d=0.2$); a small weight offers weak guidance, while a large one introduces identity-irrelevant cues such as clothing textures, harming fine-grained recognition.
These results suggest that moderate feature alignment helps the static stream learn structural priors while preserving the robustness of event representations.

\begin{table}[!t]
\centering
\caption{Ablation Studies about the number of experts in MoSE.}
\resizebox{0.95\columnwidth}{!}{
\begin{tabular}{c|cccc|cccc}
\toprule
\multirow{3}{*}{\begin{tabular}[c]{@{}c@{}}No. of Experts\\ in MoSE \end{tabular}} & \multicolumn{8}{c}{SUSTech1K-E}\\ 
\cmidrule{2-9}
  & \multicolumn{4}{c|}{\cellcolor{gray!5}\textbf{Normal Light}}  & \multicolumn{4}{c}{\cellcolor{gray!30}\textbf{Low Light}} \\
  \cmidrule{2-5} \cmidrule{6-9}
  &  NM & CL & NT & Overall &  NM & CL & NT & Overall \\ 
\midrule
1 & 86.6 & 70.3 & 78.3 & 88.4 & 74.5 & 52.8 & 61.7 & 72.5 \\
2 & 89.2 & 73.9 & 81.5 & 89.8 & 79.5 & 54.4 & 67.1 & 78.6 \\
3 & \textbf{92.5} & 78.1 & 84.8 & \textbf{92.8} & 83.8 & \textbf{58.3} & 70.5 & 83.2\\
4 & 92.4 & \textbf{78.9} & \textbf{85.1} & 92.7  & \textbf{84.2} & 58.2 & \textbf{71.1} & \textbf{83.4} \\
\bottomrule
\end{tabular}
}
\label{tab:experts}
\end{table}

\section{Conclusion and Future Work}
\label{sec:conclusion}

We present \textbf{EventGait}, an end-to-end dual-stream framework that models dynamic motion and static structural cues separately for robust event-based gait recognition. The Dynamic stream employs the Mixture of Spiking Experts (MoSE) to adaptively capture multi-timescale motion, while the Static stream utilizes Cross-modal Structure Alignment (CroSA) to transfer dense shape priors from a vision foundation model to sparse event inputs. Extensive experiments on synthesized and real-world event benchmarks demonstrate that EventGait significantly narrows the gap with camera-/LiDAR-based systems in normal lighting and substantially outperforms them under low illumination.

\noindent \textbf{Future work} will focus on: (1) Real-world scale: collecting and releasing large, diverse real-event gait datasets to reduce sim-to-real gaps. (2) Multimodal fusion: fusion with RGB/LiDAR to further boost performance.

\section{Acknowledgement}
This work was supported by the National Natural Science Foundation of China (NSFC) under Grants 62422609 and 62276243.

{
    \small
    \bibliographystyle{ieeenat_fullname}
    \bibliography{main}
}


\end{document}